\documentclass[conference]{IEEEtran}
\IEEEoverridecommandlockouts
\usepackage{cite}
\usepackage{amsmath,amssymb,amsfonts}
\usepackage{bbm}
\usepackage{algorithmic}
\usepackage{graphicx}
\usepackage{textcomp}
\usepackage{xcolor}
\def\BibTeX{{\rm B\kern-.05em{\sc i\kern-.025em b}\kern-.08em
    T\kern-.1667em\lower.7ex\hbox{E}\kern-.125emX}}
\usepackage{siunitx}
\usepackage[hidelinks]{hyperref}
\usepackage{orcidlink}

\begin{document}

\title{Towards Efficient Evaluation of Evolutionary Transfer Optimization: Case Studies on Task-Parameterized Applications
}

\author{
\IEEEauthorblockN{
Yanchen Li\textsuperscript{1}\,\orcidlink{0009-0002-2237-0451}, Xiaoming Xue\textsuperscript{2}\,\orcidlink{0000-0001-6836-7245}, and Kay Chen Tan\textsuperscript{1}\,\orcidlink{0000-0002-6802-2463}
}
\IEEEauthorblockA{
\textsuperscript{1}Department of Data Science and Artificial Intelligence, The Hong Kong Polytechnic University, Hong Kong SAR, China 
}
\IEEEauthorblockA{
\textsuperscript{2}School of Petroleum Engineering, China University of Petroleum (East China), Qingdao, China
}
\IEEEauthorblockA{
Emails:
yanchen.li@connect.polyu.hk, xuexm@upc.edu.cn, kctan@polyu.edu.hk (corresponding author: Kay Chen Tan)
}
}

\maketitle

\begin{abstract}
As evolutionary transfer optimization (ETO) scales to larger
collections of related tasks, problem evaluation can become a major
source of runtime growth.
This work studies problem-side evaluation scaling in
task-parameterized applications and reformulates application-specific
serial computations into forms suitable for parallel execution.
We organize evaluation scaling into two levels: the number of evaluated
tasks and the workload within each task.
In multi-task optimization, matrix-recursive kinematic-arm evaluation
is reformulated using an accumulation-matrix representation of cumulative
link directions.
In sequential transfer optimization, pointwise B-spline trajectory
evaluation is reformulated using a blending-matrix representation for trajectory
and collision computations.
Both reformulations maintain close numerical agreement with their
reference evaluations and substantially reduce runtime, yielding
$256.72\times$ and $93.91\times$ end-to-end speedups, respectively.
These results demonstrate problem-side reformulation as a practical
route toward scalable ETO.
Both application implementations and experimental scripts are released
as open source to support reproducibility and reuse.
\end{abstract}


\section{Introduction}

\IEEEPARstart{E}{volutionary} computation has established a versatile
population-based foundation for solving optimization problems across a
wide range of scientific and engineering domains~\cite{DBLP:journals/ec/BackS93,DBLP:journals/tec/BackHS97}.
As optimization research expanded from single-task search to settings
involving multiple related tasks, the transfer of knowledge across tasks
became a natural extension of evolutionary search.
Cross-task evolutionary research has since been explored through
representative paradigms such as multi-task optimization (MTO) and
sequential transfer optimization (STO), with notable studies spanning
both settings~\cite{DBLP:journals/tec/GuptaOF16,
DBLP:journals/tcyb/FengZZGOTQ19,
DBLP:journals/tec/XueYFZST24}.
With continued advances in cross-task optimization, evolutionary
transfer optimization (ETO) has emerged as a broader and increasingly
systematic framework for studying knowledge transfer across
optimization tasks~\cite{DBLP:journals/cim/TanFJ21,DBLP:journals/tetci/GuptaOF18,DBLP:conf/cec/Xue0FL0T25}.

As ETO has progressed toward many-task and large-scale settings,
scalability with respect to the number of tasks has received increasing
attention~\cite{DBLP:journals/cim/GuptaZOCH22, DBLP:journals/tec/HuangFQCT22,
DBLP:journals/tcyb/XueYFZST25, 11351790}.
At such scales, runtime growth can arise not only from optimization
operations, but also from the repeated evaluation of application
problems across populations and tasks.
Meanwhile, application-oriented ETO increasingly involves
task-parameterized problems, where related tasks share the same
evaluation procedure while differing in task-specific
parameters~\cite{DBLP:journals/tec/WeiLGTO26}.
In these settings, increasing the number of tasks or the internal
workload of each evaluation can make application evaluation itself a
substantial source of runtime growth.
This motivates a practical question:
\emph{Can the internal structure of some task-parameterized evaluations
be reformulated so that increasing workload does not translate directly
into increasing wall-clock time on parallel hardware?}

To characterize this scaling, we consider
task-parameterized applications through the shared evaluation form
illustrated in Fig.~\ref{fig:evaluation_scaling_formulation}.
For task $k$, the parameters $\boldsymbol{\theta}_k$ deterministically
define the task, while $\boldsymbol{\eta}$ denotes application-dependent
evaluation-workload parameters, e.g., evaluation resolution or internal sample count used within one evaluation.
The cost of evaluating a candidate solution is then repeatedly incurred over
a population of size $N$ and, when multiple tasks are evaluated, across
the task count $K$.
This characterization helps us separate two scaling regimes.
\emph{Cross-task evaluation scaling} arises as the number of
tasks $K$ increases, whereas \emph{within-task evaluation scaling}
arises when the evaluation workload increases at a fixed
task count.
In both cases, application-level computation can become
increasingly expensive when repeatedly applied throughout population-based optimization.

\begin{figure}[t]
    \centering
    \footnotesize
    \(
    \begin{aligned}
    f_k(\mathbf{x})
    &:=
    \mathcal{E}\Bigl(
        \mathbf{x};
        \underbrace{\boldsymbol{\theta}_k}_{
            \substack{
                \text{task-defining}\\
                \text{parameters}
            }
        },
        \underbrace{\boldsymbol{\eta}}_{
            \substack{
                \text{evaluation-workload}\\
                \text{parameters}
            }
        }
    \Bigr),
    \\[1.25mm]
    T_{\text{eval}}
    &\propto
    \underbrace{K}_{
        \substack{
            \text{number of}\\
            \text{tasks}
        }
    }
    \times
    \underbrace{N}_{
        \substack{
            \text{individuals}\\
            \text{per task}
        }
    }
    \times\hspace{0.5em}
    \underbrace{
        \mathcal{C}\!\left(
            \boldsymbol{\theta}_k,
            \boldsymbol{\eta}
        \right)
    }_{
        \substack{
            \text{per-individual}\\
            \text{evaluation cost}
        }
    }.
    \end{aligned}
    \)
    \caption{
        Conceptual formulation of evaluation scaling in
        task-parameterized applications.
        For task $k$, $\mathbf{x}$ denotes a candidate solution and
        $f_k(\mathbf{x})$ is evaluated through the shared evaluator
        structure $\mathcal{E}$, instantiated by the parameters 
        $\boldsymbol{\theta}_k$ and $\boldsymbol{\eta}$.
        The total evaluation time $T_{\text{eval}}$ is further
        amplified by the number of evaluated tasks $K$ and the number
        of individuals per task $N$, while
        $\mathcal{C}(\boldsymbol{\theta}_k,\boldsymbol{\eta})$ denotes
        the cost of evaluating one individual on task $k$.
    }
    \label{fig:evaluation_scaling_formulation}
    \vskip -1.9em
\end{figure}

\begin{figure*}[t!]
  \centering
  \includegraphics[width=\linewidth]{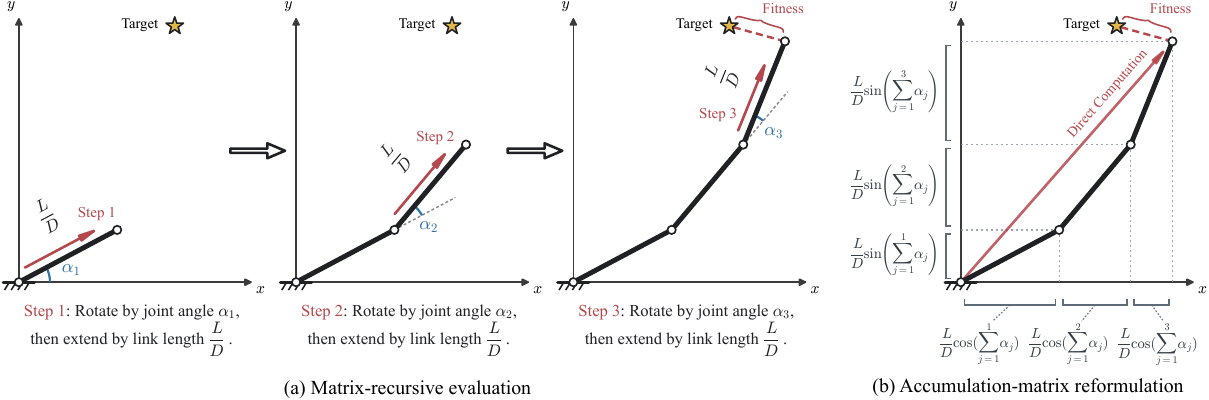}
  \vskip -0.6em
  \caption{
    Illustration and comparison of kinematic-arm evaluation for a single
    task with three joints.
    The joint angles $\alpha_1$, $\alpha_2$, and $\alpha_3$ are the
    decision variables, giving a decision dimension of $D=3$; the three
    links have equal length $L/D$, where $L$ is the total arm length.
    The target position is marked by the star, and the fitness is determined
    by the distance between the target and the end point of the final link.
    \textbf{(a)} Matrix-recursive evaluation, which computes the arm
    configuration joint by joint: at step $i$, the current orientation is
    rotated by $\alpha_i$ and then extended by one link of length $L/D$.
    \textbf{(b)} Accumulation-matrix reformulation, which obtains the link
    directions from the cumulative joint angles and
    their horizontal and vertical link projections, eliminating the
    joint-wise recursive traversal regardless of the number of joints.
  }
  \vskip -1em
\label{fig:kinematic-arm-conceptual}
\end{figure*}

These two scaling regimes suggest a practical guideline for identifying
suitable applications: evaluation cost increases under either
cross-task or within-task scaling, and the evaluation procedure
contains repeated serial calculations that can be reformulated into a
mathematically equivalent form suitable for parallel execution.
Following this guideline, this paper considers two representative
task-parameterized ETO cases:
\begin{itemize}
    \item For \emph{cross-task evaluation scaling}, we study
    kinematic-arm MTO, where the same type of evaluation is repeatedly
    performed across a large number of parameterized tasks.
    We derive an equivalent accumulation-matrix reformulation that reduces the
    serial work within each evaluation and consequently limits the runtime
    growth as the task count increases.

    \item For \emph{within-task evaluation scaling}, we study
    B-spline trajectory STO, where online optimization focuses on a
    target task while increasing the number of trajectory samples raises
    the workload within each evaluation.
    We derive an equivalent blending-matrix reformulation that reduces the
    sample-wise serial workload.
\end{itemize}
In both cases, we verify close numerical agreement between the
reformulations and their corresponding reference evaluation forms and
evaluate the runtime reduction achieved through parallel execution as
the relevant evaluation scale increases.

\section{Case Study I: Cross-Task Evaluation Scaling in Kinematic-Arm Optimization}

In multi-task optimization (MTO), multiple optimization tasks are
solved simultaneously, making the number of tasks $K$ a natural source
of evaluation scaling in ETO.
We study this cross-task scaling through kinematic-arm optimization,
where the evaluation of candidate joint configurations is repeatedly
performed across different tasks and can become increasingly costly as
$K$ grows.
We first formulate the kinematic-arm problem and its matrix-recursive
evaluation, then derive an accumulation-matrix reformulation that enables parallel
evaluation across tasks, and finally validate the reformulation through
evaluation and end-to-end MTO experiments.

\subsection{Problem Formulation and Matrix-Recursive Evaluation}

Inspired by the parametrized planar-arm setting in
\cite{DBLP:conf/gecco/MouretM20}, we consider a kinematic-arm
optimization problem in which the joint angles are optimized to bring
the end point of the final link as close as possible to a fixed target
$\mathbf{t}=[1,1]^{\mathsf T}$.
Each task $k$ is distinguished by the task parameters
$\boldsymbol{\theta}_k=[L_k,\alpha_{\text{max},k}]^{\mathsf T}$, where
$L_k$ specifies the total arm length and $\alpha_{\text{max},k}$
controls the joint-angle range.
The arm contains $D$ rotational joints and $D$ equal-length links,
while a candidate $\mathbf{x}\in[0,1]^D$ encodes the joint angles.
For task $k$, its $i$-th joint angle is
\begin{equation}
\alpha_i
:=
\frac{2\pi\alpha_{\text{max},k}}{D}
\!\cdot\!
\Bigl(x_i-\frac{1}{2}\Bigr).
\label{eq:kinematic_joint_angle}
\end{equation}
Each link consequently has length $L_k/D$.
Therefore, all tasks share the same evaluation procedure and differ only
through the task parameters $\boldsymbol{\theta}_k$.

As illustrated in Fig.~\ref{fig:kinematic-arm-conceptual}(a), the
matrix-recursive evaluation constructs the arm joint by joint.
For $i=1,\ldots,D$, step $i$ first rotates the current orientation by
$\alpha_i$ and then extends the arm by one link of length $L_k/D$.
Using planar homogeneous coordinates, this recursion is written as
\begin{equation}
\left\{
\begin{aligned}
\mathbf{M}_0
&:=
\mathbf{I}_3,
\\
\mathbf{M}_i
&:=
\mathbf{M}_{i-1}
\cdot\!
\begin{bmatrix}
\cos(\alpha_i) & -\sin(\alpha_i) & \tfrac{L_k}{D}\cos(\alpha_i)
\\[0.5ex]
\sin(\alpha_i) & \phantom{-}\cos(\alpha_i) & \tfrac{L_k}{D}\sin(\alpha_i)
\\[0.5ex]
0 & 0 & 1
\end{bmatrix},
\\
\begin{bmatrix}
\mathbf{r}_i \\
1
\end{bmatrix}
&:=
\mathbf{M}_i
\cdot
\begin{bmatrix}
0 & \!\!0\!\! & 1
\end{bmatrix}^{\mathsf T}.
\end{aligned}
\right.
\label{eq:kinematic_matrix_recursion}
\end{equation}
Here, $\mathbf{I}_3$ denotes the $3\times3$ identity matrix, and
$\mathbf{r}_i\in\mathbb{R}^2$ is the end point after the $i$-th link
is added.
Consistent with the maximization formulation adopted in previous
kinematic-arm work~\cite{DBLP:conf/gecco/MouretM20}, the fitness of
task $k$ is defined as
\begin{equation}
f_k(\mathbf{x})
:=
-\left\|
\mathbf{r}_D-\mathbf{t}
\right\|_2.
\label{eq:kinematic_fitness}
\end{equation}
Since each $\mathbf{M}_i$ depends on $\mathbf{M}_{i-1}$, the
evaluation follows a sequential path over the $D$ joints.
This joint-wise recursion is repeated for every individual in a
population, so its evaluation cost is first amplified by the population
size $N$ and further amplified in MTO as the number of tasks
$K$ increases.

\begin{figure*}[t!]
  \centering
  \includegraphics[width=\linewidth]{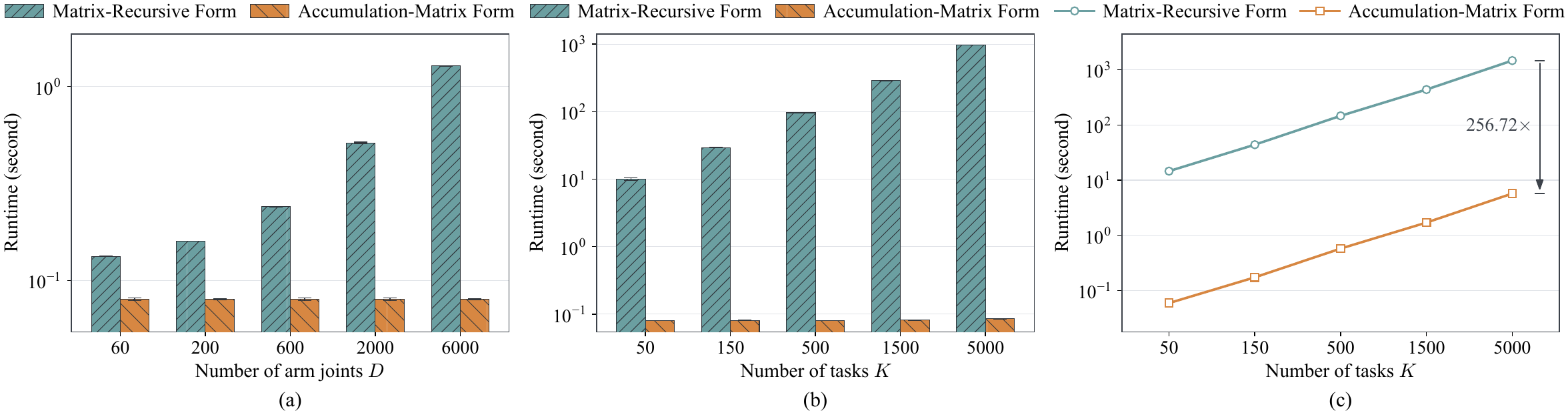}
  \vskip -0.8em
  \caption{
    Evaluation and end-to-end runtime results for the kinematic-arm optimization case study.
    All runtime values are reported as the mean over three repeated measurements, with error bars indicating the standard deviation.
    \textbf{(a)} Evaluation runtime for one individual under increasing arm joint counts $D$, comparing the matrix-recursive evaluation with its accumulation-matrix reformulation.
    \textbf{(b)} Evaluation runtime under increasing task counts $K$, where $K$ denotes the number of evaluated tasks, comparing the same two forms at a fixed joint count.
    \textbf{(c)} One-generation end-to-end runtime of MA-MTO under increasing $K$; the annotated multiplier denotes the runtime speedup obtained by the accumulation-matrix reformulation over the matrix-recursive evaluation at the largest tested task count.
  }
  \vskip -1em
\label{fig:kinematic-arm-results}
\end{figure*}

\subsection{Accumulation-Matrix Reformulation}

The matrix-recursive evaluation requires each transformation
$\mathbf{M}_i$ to depend on $\mathbf{M}_{i-1}$. However, the absolute
orientation of the $i$-th link is simply the cumulative angle
$\alpha_1+\cdots+\alpha_i$. Therefore, the recursive transformation
chain can be removed by computing all cumulative link directions
jointly, as illustrated in
Fig.~\ref{fig:kinematic-arm-conceptual}(b).

Let
$\boldsymbol{\alpha}
=[\alpha_1,\ldots,\alpha_D]^{\mathsf T}$
collect the relative joint angles and
$\boldsymbol{\phi}
=[\phi_1,\ldots,\phi_D]^{\mathsf T}$
collect the resulting link directions.
Their relation can be expressed by the fixed prefix-sum matrix
\begin{equation}
\underbrace{
\begin{bmatrix}
\phi_1\\
\phi_2\\
\vdots\\
\phi_D
\end{bmatrix}
}_{\boldsymbol{\phi}}
:=
\begin{bmatrix}
1 & 0 & \cdots & 0\\[-1ex]
1 & 1 & \ddots & \vdots\\[-1ex]
\vdots & \vdots & \ddots & 0\\
1 & 1 & \cdots & 1
\end{bmatrix}
\!\cdot\!
\underbrace{
\begin{bmatrix}
\alpha_1\\
\alpha_2\\
\vdots\\
\alpha_D
\end{bmatrix}
}_{\boldsymbol{\alpha}}.
\label{eq:kinematic_cumulative_angles}
\end{equation}
This replaces the joint-wise transformation recursion in
Eq.~\eqref{eq:kinematic_matrix_recursion} with a bulk cumulative-sum
operation.

Once the cumulative directions are obtained, each link contributes
directly through its horizontal and vertical projections. The final
end point is therefore computed as
\begin{equation}
\mathbf{r}_D
:=
\frac{L_k}{D}
\!\cdot\!
\begin{bmatrix}
\displaystyle\sum_{i=1}^{D}\cos(\phi_i)
&
\displaystyle\sum_{i=1}^{D}\sin(\phi_i)
\end{bmatrix}^{\mathsf T}.
\label{eq:kinematic_direct_position}
\end{equation}
This expression directly produces the same final position used in
Eq.~\eqref{eq:kinematic_fitness}, without constructing the intermediate
transformation matrices $\mathbf{M}_1,\ldots,\mathbf{M}_D$.
The cumulative directions, trigonometric projections, and reductions
can consequently be performed as joint operations that are well suited
to modern parallel processors. For population-based MTO, additional
candidate and task axes can be introduced directly to the same
operations, allowing the $N$ candidates across $K$ tasks to be
evaluated jointly without restoring the joint-wise recursion.

\subsection{Experimental Validation}

We construct a reproducible task-parameterized kinematic-arm testbed,
where task parameters $[L_k,\alpha_{\mathrm{max},k}]^{\mathsf T}$ are
generated using a scrambled Sobol sequence~\cite{SOBOL196786,OWEN1998466}.
All experiments are conducted on a single NVIDIA GeForce RTX 3090 GPU
with an Intel(R) Xeon(R) Platinum 8350C CPU @ 2.60\,GHz.
We first verify the numerical agreement between the accumulation-matrix
reformulation and the matrix-recursive evaluation using identical
candidate solutions.
Across 10 random seeds with $K=5000$, $N=16$, and $D=60$, the two
evaluation forms achieved a mean absolute fitness discrepancy of
$2.561\times10^{-8}\pm1.600\times10^{-10}$, where each seed-level
value equally averages the task-wise discrepancies over all 5000
tasks. The maximum absolute discrepancy over all seeds, tasks, and
individuals was $3.576\times10^{-7}$.

We then isolate the two primary scaling factors in
Fig.~\ref{fig:kinematic-arm-results}(a) and
Fig.~\ref{fig:kinematic-arm-results}(b).
Fig.~\ref{fig:kinematic-arm-results}(a) evaluates one individual on
one task while increasing the number of joints $D$.
The runtime of the matrix-recursive evaluation grows markedly with
$D$, whereas the accumulation-matrix reformulation remains low and nearly constant
over the tested range.
Fig.~\ref{fig:kinematic-arm-results}(b) fixes $N=16$ and $D=60$
while increasing the number of evaluated tasks $K$.
A similar trend is observed across tasks: the matrix-recursive
evaluation scales rapidly with $K$, while the accumulation-matrix reformulation
shows little runtime growth over the tested task counts, demonstrating
the benefit of exposing the evaluation to parallel execution.

For the end-to-end validation, we use mean alignment multi-task
optimization (MA-MTO), which adopts a mean-aligned elite-injection
strategy inspired by previous transfer work~\cite{DBLP:journals/tec/LiZTZ22}
and is implemented with EvoX~\cite{DBLP:journals/tec/HuangCLJT25}
for population-level parallel computation while retaining task-wise
execution.
Each task evolves through GA-based evolution, while transferred elites
are aligned to the mean of the receiving population before injection.
As shown in Fig.~\ref{fig:kinematic-arm-results}(c), the runtime
advantage of the accumulation-matrix reformulation is retained within the complete
optimization process, reaching a $256.72\times$ speedup at $K=5000$.
Across 10 independent runs with $K=5000$, $N=16$, $D=60$, and a
100-generation budget, MA-MTO achieved a median final fitness of
$-0.985\pm3.368\times10^{-4}$ and a 95th-percentile final fitness of
$-0.541\pm3.948\times10^{-4}$.
For each run, the median and 95th percentile are computed across the
5000 tasks, while the reported mean and standard deviation summarize
the 10 independent runs.

\section{Case Study II: Within-Task Evaluation Scaling in B-Spline Trajectory Optimization}

In sequential transfer optimization (STO), optimization of a target
task reuses knowledge accumulated from previously solved source tasks,
while online evaluation primarily focuses on the current target task.
Nevertheless, a single target task can still incur substantial
evaluation cost when the application requires an increasingly high
evaluation resolution. We study this setting through trajectory
optimization parameterized by a B-spline, i.e., a smooth trajectory
curve whose shape is controlled by a compact set of control points and
whose evaluation resolution can be increased independently of the
decision dimension. We first formulate the trajectory problem and its
direct pointwise evaluation, then derive a blending-matrix reformulation that
enables parallel computation, and finally validate the reformulation
through standalone evaluation and end-to-end STO experiments.

\subsection{Problem Formulation and Pointwise Evaluation}

We consider a continuous-variable trajectory optimization problem in a
normalized two-dimensional workspace. For task $k$, the start and goal
positions are fixed at $(0,0)$ and $(1,1)$, respectively, while the
task is distinguished by $M$ axis-aligned square obstacles
$\{\mathcal{O}_{k,m}\}_{m=1}^{M}$. The objective is to generate a short
trajectory from the start to the goal without intersecting these
obstacles. A candidate solution
$\mathbf{y}=[y_1,\ldots,y_D]^{\mathsf T}\in[0,1]^D$ determines the
vertical coordinates of $D$ internal control points, while their
horizontal coordinates are uniformly fixed. Together with the two
endpoints, the $C=D+2$ control points are
\begin{equation}
\mathbf{p}_i :=
\begin{cases}
[0,0]^{\mathsf T}, & i=0,\\[1mm]
\left[i/(D+1),\,y_i\right]^{\mathsf T}, & 1\le i\le D,\\[1mm]
[1,1]^{\mathsf T}, & i=D+1.
\end{cases}
\label{eq:trajectory_control_points}
\end{equation}

A B-spline forms a smooth trajectory by assigning
position-dependent coefficients to neighboring control points and
combining them according to these coefficients.
We use a cubic B-spline with degree $\delta=3$.
For a normalized curve parameter ranging from $0$ at the start to $1$
at the goal, the coefficient $b_{i,\delta}(\cdot)$ specifies the
contribution of control point $\mathbf{p}_i$ and is computed using the
Cox--de Boor recursion~\cite{10.1093/imamat/10.2.134,DEBOOR197250}.
The recursion uses a fixed sequence over $[0,1]$: the two boundary
values $0$ and $1$ are each repeated $\delta+1$ times, while the
intermediate values are uniformly placed at $j/(C-\delta)$ for
$j=1,\ldots,C-\delta-1$. This configuration makes the resulting curve
pass through the prescribed start and goal points.

Using $R$ equally spaced sampling positions along the trajectory, the pointwise evaluation
moves a local window along the control points and forms each trajectory
sample as
\begin{equation}
\mathbf{q}_r :=
\begin{cases}
\displaystyle
\sum_{i=
\bigl\lfloor\!
\frac{r(C-\delta)}{R-1}
\!\bigr\rfloor
}^{\bigl\lfloor\!
\frac{r(C-\delta)}{R-1}
\!\bigr\rfloor+\delta}\!
b_{i,\delta}\!\Bigl(\frac{r}{R-1}\Bigr)\cdot
\mathbf{p}_i,
&
0\le r\le R-2,
\\[3mm]
\mathbf{p}_{C-1},
&
r=R-1.
\end{cases}
\label{eq:trajectory_pointwise_sampling}
\end{equation}
Here, $R$ instantiates the evaluation-workload parameter
$\boldsymbol{\eta}$ introduced in
Fig.~\ref{fig:evaluation_scaling_formulation}: increasing $R$ evaluates
the same candidate using more trajectory samples without changing the
decision dimension $D$.
As $r$ increases, the summation bounds move the local window along the
control points. Each sampling position therefore combines only
$\delta+1=4$ neighboring control points in the cubic case.
The pointwise evaluation performs these local combinations sequentially
over the $R$ trajectory samples.

Since the objective is to obtain a short trajectory while avoiding
obstacles, its fitness combines the sampled path length with collision
penalties. We regard $\mathcal{O}_{k,m}$ as the closed square region
occupied by the $m$-th obstacle of task $k$. Accordingly, the fitness
is defined by
\begin{equation}
\begin{aligned}
f_k(\mathbf{y})
:={}&
\sum_{r=0}^{R-2}
\left\|
\mathbf{q}_{r+1}-\mathbf{q}_r
\right\|_2
\\
&+
\rho\cdot
\sum_{m=1}^{M}
\mathbbm{1}\!\left[
\exists\,r:
\overline{\mathbf{q}_r\mathbf{q}_{r+1}}
\cap
\mathcal{O}_{k,m}
\neq
\emptyset
\right].
\end{aligned}
\label{eq:trajectory_fitness}
\end{equation}
Here, $\rho$ denotes the penalty assigned to each collided obstacle,
and $\mathbbm{1}[\cdot]$ denotes the indicator function.
The line segment
$\overline{\mathbf{q}_r\mathbf{q}_{r+1}}$ connects two consecutive
trajectory samples. An obstacle contributes the penalty once if any
trajectory segment intersects its square region, and boundary contact
is also treated as an intersection.

The pointwise evaluation follows this construction directly.
For each candidate, the $R$ trajectory samples are formed sequentially,
and each newly formed segment is tested jointly against all $M$
obstacles. Although the local control-point combination and the
obstacle checks at each sampling position are already performed
jointly, traversal over the $R$ trajectory samples remains sequential.
Consequently, increasing $R$ enlarges the serial within-task evaluation
workload and can lead to substantially higher evaluation time.

\subsection{Blending-Matrix Reformulation}

The pointwise evaluation retains a sequential traversal over the $R$
trajectory samples because each sampling position operates on a
different local window of control points. To remove this traversal, we
first align the local coefficients of all sampling positions to a common
control-point axis. Specifically, the coefficient matrix
$\mathbf{B}\!\in\!\mathbb{R}^{R\times C}$ is defined element-wise as
\begin{equation}
\mathbf{B}_{r,i}
:=
\begin{cases}
\displaystyle
b_{i,\delta}\!\biggl(\frac{r}{R-1}\biggr),
&
\begin{aligned}
\Bigl\lfloor\!\tfrac{r(C-\delta)}{R-1}\!\Bigr\rfloor &\le i\le \Bigl\lfloor\!\tfrac{r(C-\delta)}{R-1}\!\Bigr\rfloor\!\!+\!\delta
\\[-0.5mm]
0 &\le r\le R-2,
\end{aligned}
\\[5mm]
1,
&
i=C-1,\; r=R-1,
\\[1mm]
0,
&
\text{otherwise}.
\end{cases}
\label{eq:trajectory_coefficient_matrix}
\end{equation}
Each of the first $R-1$ rows therefore places the $\delta+1$ coefficients
of one local control-point window into their corresponding columns, with
all remaining entries set to zero. The final row selects the prescribed
goal point. In this way, the moving local windows of the pointwise
evaluation are represented within a common matrix structure.

\begin{figure*}[t!]
  \centering
  \includegraphics[width=0.97\linewidth]{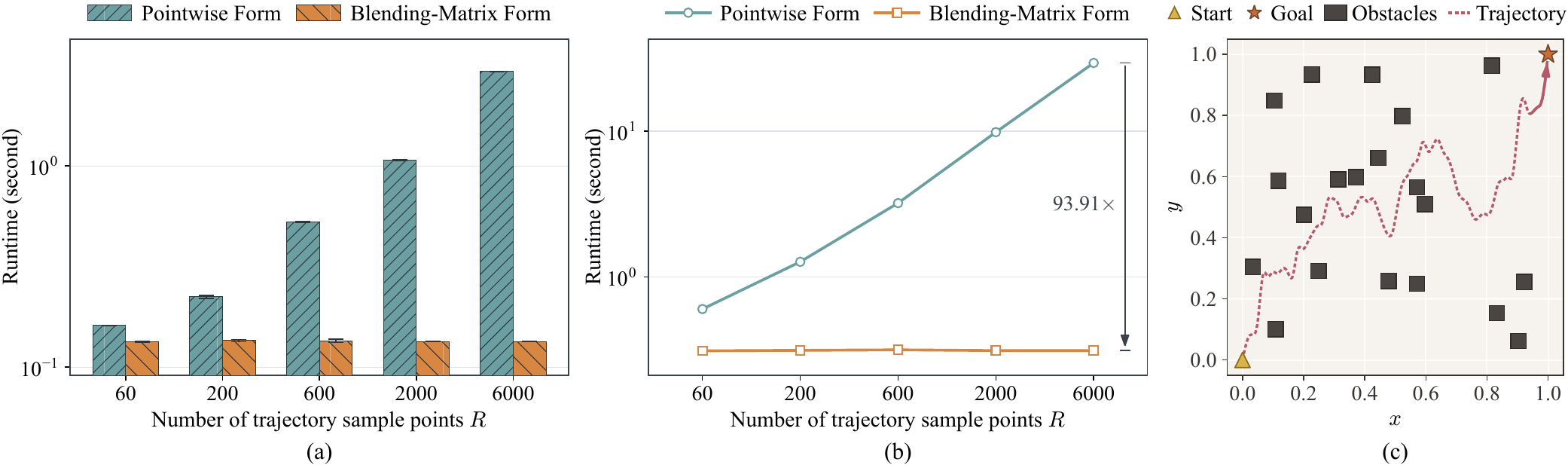}
  \vskip -0.8em
  \caption{
    Evaluation and end-to-end results for the B-spline trajectory optimization case study.
    Runtime values in \textbf{(a)} and \textbf{(b)} are reported as the mean over three repeated measurements, with error bars indicating the standard deviation.
    \textbf{(a)} Evaluation runtime under increasing numbers of trajectory sample points $R$, where $R$ denotes the number of points used to discretize each B-spline trajectory for evaluation, comparing the pointwise evaluation with its blending-matrix reformulation.
    \textbf{(b)} One-generation end-to-end runtime of MS-STO under increasing $R$; the annotated multiplier denotes the runtime speedup obtained by the blending-matrix reformulation over the pointwise evaluation at the largest tested $R$.
    \textbf{(c)} Best feasible trajectory obtained over ten independent MS-STO runs under the fixed 100-generation budget, where $x$ and $y$ denote the normalized horizontal and vertical spatial coordinates, respectively.
  }
  \vskip -1em
\label{fig:trajectory-optimization-results}
\end{figure*}

Stacking the control points and trajectory samples by rows then allows
the complete trajectory to be obtained through a single matrix
multiplication, schematically written as
\begin{equation}
\underbrace{
\begin{bmatrix}
\mathbf{q}_0^{\mathsf T}\\[1ex]
\vdots\\[0.5ex]
\mathbf{q}_{R-2}^{\mathsf T}\\[1ex]
\mathbf{q}_{R-1}^{\mathsf T}
\end{bmatrix}
}_{\mathbf{Q}}
\!:=\!
\underbrace{
\begin{bmatrix}
b_{0,\delta} & \!\!\cdots\!\!     & b_{\delta,\delta}     & \!\!\mathbf{0}\!\! & 0              \\[0ex]
\mathbf{0}   & \!\!\ddots\!\!     & \vdots                & \!\!\ddots\!\!     & \mathbf{0}     \\[1ex]
0            & \!\!\cdots\!\!     & b_{C-\delta-1,\delta} & \!\!\cdots\!\!     & b_{C-1,\delta} \\[1ex]
0            & \!\!\mathbf{0}\!\! & 0                     & \!\!\mathbf{0}\!\! & 1
\end{bmatrix}
}_{\mathbf{B}}
\!\cdot\!
\underbrace{
\begin{bmatrix}
\mathbf{p}_0^{\mathsf T}\\[1ex]
\vdots\\[1ex]
\mathbf{p}_{C-1}^{\mathsf T}
\end{bmatrix}
}_{\mathbf{P}}.
\label{eq:trajectory_matrix_sampling}
\end{equation}
For readability, the sampling-position arguments are omitted from the
schematic coefficients; their exact values and positions are given by
Eq.~\eqref{eq:trajectory_coefficient_matrix}. Each of the first $R-1$
rows contains only $\delta+1$ consecutive nonzero coefficients, while
$\mathbf{0}$ denotes an all-zero block. Compared with
processing the local windows sample by sample, this form replaces the
sequential traversal over the $R$ sampling positions with a unified
matrix computation that is well suited to modern parallel processors.

The path-length term in Eq.~\eqref{eq:trajectory_fitness} can also
be computed from the complete trajectory matrix $\mathbf{Q}$ without a
pointwise traversal. All consecutive-point differences are formed
jointly through
\begin{equation}
\begin{bmatrix}
(\mathbf{q}_1-\mathbf{q}_0)^{\mathsf T} \\[0.25ex]
(\mathbf{q}_2-\mathbf{q}_1)^{\mathsf T} \\[0.25ex]
\vdots \\[0.25ex]
(\mathbf{q}_{R-1}-\mathbf{q}_{R-2})^{\mathsf T}
\end{bmatrix}
\!=\!
\begin{bmatrix}
-1                & \phantom{-}1 & 0      & \cdots & 0      \\[-0.5ex]
\phantom{-}0      & -1           & 1      & \ddots & \vdots \\[-0.75ex]
\phantom{-}\vdots & \ddots       & \ddots & \ddots & 0      \\[0ex]
\phantom{-}0      & \cdots       & 0      & -1       & 1
\end{bmatrix}
\!\cdot
\mathbf{Q}
.
\label{eq:trajectory_segment_matrix}
\end{equation}
The Euclidean norms of the resulting rows are then computed jointly and
summed to obtain the complete path length.

For collision evaluation, the pointwise form already evaluates the
current trajectory segment against all $M$ obstacles jointly. Its
sequential execution therefore arises from traversing the $R-1$
trajectory segments. We remove this traversal by organizing all
segment--obstacle relations into
\begin{equation}
\mathbf{H}_k
:=
\underset{
0\le r\le R-2,\;1\le m\le M
}{
\Bigl[
\mathbbm{1}\!\bigl[
\overline{\mathbf{Q}_{r,:}\mathbf{Q}_{r+1,:}}
\cap
\mathcal{O}_{k,m}
\neq
\emptyset
\bigr]
\Bigr]
}
\!\in\!
\{0,1\}^{(R-1)\times M}.
\label{eq:trajectory_collision_matrix}
\end{equation}
This form evaluates the segment--obstacle relations for all $R-1$
trajectory segments jointly while preserving the existing joint
computation over the $M$ obstacles. A column-wise logical reduction then
determines whether each obstacle is intersected by any trajectory
segment, yielding the collision term in
Eq.~\eqref{eq:trajectory_fitness}.

Although the reformulation above is described for one candidate on one
task, its matrix structure extends naturally to population-based
evaluation. Multiple candidates can be handled jointly by introducing
an additional candidate axis to the same matrix operations. 
These extensions are natural for modern parallel
processors and do not change the underlying evaluation formulation.
Consequently, reducing the sequential workload of a single evaluation
is particularly important because this cost is repeatedly amplified by
the population size and, where applicable, by the number of evaluated
tasks.

\subsection{Experimental Validation}

Inspired by the B-spline trajectory optimization settings in prior
evolutionary transfer optimization studies~\cite{DBLP:conf/ieeecai/LinLXT24,DBLP:journals/tec/LiZTZ22},
we construct a reproducible task-parameterized trajectory optimization
testbed with explicitly specified spline, obstacle, and evaluation
configurations. Each task uses $D=60$ decision variables and $M=20$
square obstacles of side length $0.05$, with the collision penalty set
to $\rho=20$. To examine within-task evaluation scaling independently
of the decision dimension, we vary the number of trajectory sample
points as
$R\in\{60,200,600,2000,6000\}$ while keeping $D$ fixed.
For reproducibility and to avoid trivial or infeasible layouts, obstacle centers
are generated using a scrambled Sobol sequence~\cite{SOBOL196786,OWEN1998466}
and retained only when the obstacles stay away from the start and goal,
remain separated from one another, block the straight start--goal path,
and still leave at least one collision-free route verified by a simple
grid search.
All experiments are conducted on a single NVIDIA GeForce RTX 3090 GPU
with an Intel(R) Xeon(R) Platinum 8350C CPU @ 2.60\,GHz.

We first verify the numerical agreement between the matrix
reformulation and the pointwise evaluation using identical candidate
solutions.
Across 10 random seeds with $K=1$ task, population size
$N=16$, $D=60$, and $R=6000$, the pointwise evaluation and its matrix
reformulation achieved a mean absolute fitness discrepancy of
$1.101\times10^{-5}\pm3.011\times10^{-6}$, where each seed-level
value averages the 16 paired individual discrepancies.
The maximum absolute discrepancy over all seeds and individuals was
$3.052\times10^{-5}$.
We then isolate the effect of $R$ through single-individual evaluation
on a single task, as shown in
Fig.~\ref{fig:trajectory-optimization-results}(a).
The runtime of the pointwise evaluation increases markedly with $R$,
whereas the matrix reformulation remains low and nearly constant over
the tested range.
This behavior reflects the parallel execution exposed along the
sampling-point dimension: while sufficient parallel processing
capacity is available, increasing $R$ need not translate into a
comparable increase in wall-clock time.

For the end-to-end validation, we adopt mean similarity sequential
transfer optimization (MS-STO)~\cite{DBLP:journals/tec/XueYFZST24},
which reuses source solutions according to their mean-based similarity
to the target population and is implemented with
EvoX~\cite{DBLP:journals/tec/HuangCLJT25} for population-level parallel
computation.
GA-based evolution is used for both offline source-knowledge
construction and online target optimization.
The experiment uses a target population size of $N=16$ and a knowledge
database constructed from $K=5000$ previously solved source tasks; the
offline source construction is excluded from the reported online runtime.
As shown in Fig.~\ref{fig:trajectory-optimization-results}(b), the
pointwise end-to-end runtime grows rapidly with $R$, while the matrix
reformulation remains nearly unchanged, yielding a $93.91\times$
speedup at $R=6000$.
Across 10 independent runs with $D=60$, $R=6000$, and a
100-generation budget, MS-STO obtained collision-free trajectories in
$4/10$ runs.
Among the feasible runs, the final trajectory length was
$3.365\pm0.587$, with a best feasible length of $2.586$, as illustrated
in Fig.~\ref{fig:trajectory-optimization-results}(c).

\section{Conclusion}

This paper studied problem-side evaluation scaling in
task-parameterized evolutionary transfer optimization.
The two case studies have exposed two key scaling regimes:
cross-task scaling in kinematic-arm optimization and within-task
scaling in B-spline trajectory optimization.
In both cases, mathematically equivalent reformulations replace
recursive or sample-wise serial computation with structured matrix
forms suitable for parallel execution, while maintaining close numerical
agreement and substantially reducing isolated evaluation and
end-to-end runtime.
The results have shown that application evaluation itself provides a
practical opportunity for improving ETO efficiency at increasing task
scales.

The present study is limited to two continuous task-parameterized
applications, and the reformulations remain application-specific.
Some matrix forms may also trade additional intermediate storage for
lower runtime, making memory another consideration at larger scales.
Future work should identify reusable evaluation-reformulation
principles across broader problem classes and further address
algorithm-side scaling in transfer, variation, selection, and other
optimization operations.
Combining efficient problem evaluation with scalable algorithmic
execution is supposed to ultimately enable ETO systems to handle substantially
larger collections of tasks.

\let\hyperurl\url
\renewcommand{\url}[1]{\href{#1}{#1}}

\section*{Code Availability}
The implementations of both application case studies and all experimental
scripts used in this paper are publicly available at
\url{https://github.com/liyc5929/task-parameterized-eto-evaluation}.

\let\url\hyperurl

\bibliographystyle{IEEEtran}
\bibliography{main}

@article{DBLP:journals/ec/BackS93,
  author       = {Thomas B{\"{a}}ck and
                  Hans{-}Paul Schwefel},
  title        = {An overview of evolutionary algorithms for parameter optimization},
  journal      = {Evol. Comput.},
  volume       = {1},
  number       = {1},
  pages        = {1--23},
  year         = {1993},
}

@article{DBLP:journals/tec/BackHS97,
  author       = {Thomas B{\"{a}}ck and
                  Ulrich Hammel and
                  Hans{-}Paul Schwefel},
  title        = {Evolutionary computation: {Comments} on the history and current state},
  journal      = {{IEEE} Trans. Evol. Comput.},
  volume       = {1},
  number       = {1},
  pages        = {3--17},
  year         = {1997},
}

@article{DBLP:journals/tec/GuptaOF16,
  author       = {Abhishek Gupta and
                  Yew{-}Soon Ong and
                  Liang Feng},
  title        = {Multifactorial evolution: {Toward} evolutionary multitasking},
  journal      = {{IEEE} Trans. Evol. Comput.},
  volume       = {20},
  number       = {3},
  pages        = {343--357},
  year         = {2016},
}

@article{DBLP:journals/tcyb/FengZZGOTQ19,
  author       = {Liang Feng and
                  Lei Zhou and
                  Jinghui Zhong and
                  Abhishek Gupta and
                  Yew{-}Soon Ong and
                  Kay Chen Tan and
                  A. K. Qin},
  title        = {Evolutionary multitasking via explicit autoencoding},
  journal      = {{IEEE} Trans. Cybern.},
  volume       = {49},
  number       = {9},
  pages        = {3457--3470},
  year         = {2019},
}

@Article{DBLP:journals/cim/TanFJ21,
  author  = {Kay Chen Tan and Liang Feng and Min Jiang},
  journal = {{IEEE} Comput. Intell. Mag.},
  title   = {Evolutionary transfer optimization - {A} new frontier in evolutionary computation research},
  year    = {2021},
  number  = {1},
  pages   = {22--33},
  volume  = {16},
  doi     = {10.1109/MCI.2020.3039066},
}

@article{DBLP:journals/tetci/GuptaOF18,
  author       = {Abhishek Gupta and
                  Yew{-}Soon Ong and
                  Liang Feng},
  title        = {Insights on transfer optimization: {Because} experience is the best
                  teacher},
  journal      = {{IEEE} Trans. Emerg. Top. Comput. Intell.},
  volume       = {2},
  number       = {1},
  pages        = {51--64},
  year         = {2018},
}

@inproceedings{DBLP:conf/cec/Xue0FL0T25,
  author       = {Xiaoming Xue and
                  Liang Feng and
                  Yinglan Feng and
                  Rui Liu and
                  Kai Zhang and
                  Kay Chen Tan},
  title        = {A theoretical analysis of analogy-based evolutionary transfer optimization},
  booktitle    = {{IEEE} Congress on Evolutionary Computation, {CEC}},
  pages        = {1--8},
  publisher    = {{IEEE}},
  year         = {2025},
}

@Article{DBLP:journals/tec/HuangFQCT22,
  author  = {Yuxiao Huang and Liang Feng and Alex Kai Qin and Meng Chen and Kay Chen Tan},
  journal = {{IEEE} Trans. Evol. Comput.},
  title   = {Toward large-scale evolutionary multitasking: {A} {GPU}-based paradigm},
  year    = {2022},
  number  = {3},
  pages   = {585--598},
  volume  = {26},
  doi     = {10.1109/TEVC.2021.3110506},
}

@article{DBLP:journals/cim/GuptaZOCH22,
  author       = {Abhishek Gupta and
                  Lei Zhou and
                  Yew{-}Soon Ong and
                  Zefeng Chen and
                  Yaqing Hou},
  title        = {Half a dozen real-world applications of evolutionary multitasking,
                  and more},
  journal      = {{IEEE} Comput. Intell. Mag.},
  volume       = {17},
  number       = {2},
  pages        = {49--66},
  year         = {2022},
}

@INPROCEEDINGS{11351790,
  author={Zhao, Jiawei and Chen, Xuefeng and Feng, Liang},
  booktitle={International Conference on Machine Intelligence and Nature-Inspired Computing (MIND)}, 
  title={An efficient index-based source task selection approach for large-scale evolutionary sequential transfer optimization}, 
  year={2025},
  volume={},
  number={},
  pages={229-230},
}

@article{DBLP:journals/tec/XueYFZST24,
  author       = {Xiaoming Xue and
                  Cuie Yang and
                  Liang Feng and
                  Kai Zhang and
                  Linqi Song and
                  Kay Chen Tan},
  title        = {Solution transfer in evolutionary optimization: {An} empirical study
                  on sequential transfer},
  journal      = {{IEEE} Trans. Evol. Comput.},
  volume       = {28},
  number       = {6},
  pages        = {1776--1793},
  year         = {2024},
}

@article{DBLP:journals/tec/WeiLGTO26,
  author       = {Tingyang Wei and
                  Jiao Liu and
                  Abhishek Gupta and
                  Puay Siew Tan and
                  Yew{-}Soon Ong},
  title        = {\(({\theta}_{\text{l}}\),\({\theta}_{\text{u}})\)-parametric
                  multitask optimization: {Joint} search in solution and infinite task
                  spaces},
  journal      = {{IEEE} Trans. Evol. Comput.},
  volume       = {30},
  number       = {3},
  pages        = {1270--1283},
  year         = {2026},
}

@article{DBLP:journals/tcyb/XueYFZST25,
  author       = {Xiaoming Xue and
                  Cuie Yang and
                  Liang Feng and
                  Kai Zhang and
                  Linqi Song and
                  Kay Chen Tan},
  title        = {A scalable test problem generator for sequential transfer optimization},
  journal      = {{IEEE} Trans. Cybern.},
  volume       = {55},
  number       = {5},
  pages        = {2110--2123},
  year         = {2025},
}

@inproceedings{DBLP:conf/gecco/MouretM20,
  author       = {Jean{-}Baptiste Mouret and
                  Glenn Maguire},
  title        = {Quality diversity for multi-task optimization},
  booktitle    = {Genetic and Evolutionary Computation Conference, {GECCO}},
  pages        = {121--129},
  publisher    = {{ACM}},
  year         = {2020},
}

@inproceedings{DBLP:conf/ieeecai/LinLXT24,
  author       = {Wu Lin and
                  Qiuzhen Lin and
                  Xiaoming Xue and
                  Kay Chen Tan},
  title        = {Sequential transfer via clustering-based similarity measurement for
                  faster trajectory optimization},
  booktitle    = {{IEEE} Conference on Artificial Intelligence, {CAI}},
  pages        = {1296--1301},
  publisher    = {{IEEE}},
  year         = {2024},
}

@article{DBLP:journals/tec/LiZTZ22,
  author       = {Jian{-}Yu Li and
                  Zhi{-}Hui Zhan and
                  Kay Chen Tan and
                  Jun Zhang},
  title        = {A meta-knowledge transfer-based differential evolution for multitask
                  optimization},
  journal      = {{IEEE} Trans. Evol. Comput.},
  volume       = {26},
  number       = {4},
  pages        = {719--734},
  year         = {2022},
}

@article{10.1093/imamat/10.2.134,
    author = {COX, M. G.},
    title = {The numerical evaluation of {B}-splines},
    journal = {IMA Journal of Applied Mathematics},
    volume = {10},
    number = {2},
    pages = {134-149},
    year = {1972},
    month = {10},
    issn = {0272-4960},
}

@article{DEBOOR197250,
title = {On calculating with {B}-splines},
journal = {Journal of Approximation Theory},
volume = {6},
number = {1},
pages = {50-62},
year = {1972},
issn = {0021-9045},
author = {Carl {de Boor}}
}

@article{SOBOL196786,
title = {On the distribution of points in a cube and the approximate evaluation of integrals},
journal = {USSR Computational Mathematics and Mathematical Physics},
volume = {7},
number = {4},
pages = {86-112},
year = {1967},
issn = {0041-5553},
author = {I.M Sobol'}
}

@article{OWEN1998466,
title = {Scrambling {Sobol'} and {Niederreiter--Xing} points},
journal = {Journal of Complexity},
volume = {14},
number = {4},
pages = {466-489},
year = {1998},
issn = {0885-064X},
author = {Art B. Owen},
}

@article{DBLP:journals/tec/HuangCLJT25,
  author       = {Beichen Huang and
                  Ran Cheng and
                  Zhuozhao Li and
                  Yaochu Jin and
                  Kay Chen Tan},
  title        = {{EvoX}: {A} distributed {GPU}-accelerated framework for scalable evolutionary
                  computation},
  journal      = {{IEEE} Trans. Evol. Comput.},
  volume       = {29},
  number       = {5},
  pages        = {1649--1662},
  year         = {2025},
}

\end{document}